\pdfoutput=1

\documentclass[runningheads]{llncs}
\usepackage{graphicx}
\usepackage{amssymb}
\usepackage{multirow}
\usepackage{graphicx}
\begin{document}

\title{Spatially Localized Atlas Network Tiles \\ 
Enables 3D Whole Brain Segmentation \\ 
from Limited Data}
%
%
\author{Yuankai Huo*\inst{1} \and
Zhoubing Xu\inst{1} \and
Katherine Aboud\inst{1} \and
Prasanna Parvathaneni\inst{1} \and
Shunxing Bao\inst{1} \and
Camilo Bermudez\inst{1} \and
Susan M. Resnick\inst{2} \and
Laurie E. Cutting\inst{1} \and
Bennett A. Landman\inst{1}} 
\authorrunning{Y. Huo et al.}
\titlerunning{SLANT 3D Whole Brain Segmentation from Limited Data}
%
\institute{Vanderbilt University, Nashville, TN \and
Laboratory of Behavioral Neuroscience, National Institute on Aging, MD}
\maketitle              
\begin{abstract}
Whole brain segmentation on a structural magnetic resonance imaging (MRI) is essential in non-invasive investigation for neuroanatomy. Historically, multi-atlas segmentation (MAS) has been regarded as the \textit{de facto} standard method for whole brain segmentation. Recently, deep neural network approaches have been applied to whole brain segmentation by learning random patches or 2D slices. Yet, few previous efforts have been made on detailed whole brain segmentation using 3D networks due to the following challenges: (1) fitting entire whole brain volume into 3D networks is restricted by the current GPU memory, and (2) the large number of targeting labels (e.g., $>$ 100 labels) with limited number of training 3D volumes (e.g., $<$ 50 scans). In this paper, we propose the spatially localized atlas network tiles (SLANT) method to distribute multiple independent 3D fully convolutional networks to cover overlapped sub-spaces in a standard atlas space. This strategy simplifies the whole brain learning task to localized sub-tasks, which was enabled by combing canonical registration and label fusion techniques with deep learning. To address the second challenge, auxiliary labels on 5111 initially unlabeled scans were created by MAS for pre-training. From empirical validation, the state-of-the-art MAS method achieved mean Dice value of 0.76, 0.71, and 0.68, while the proposed method achieved 0.78, 0.73, and 0.71 on three validation cohorts. Moreover, the computational time reduced from $>$ 30 hours using MAS to $\approx$ 15 minutes using the proposed method. The source code is available online \footnote{https://github.com/MASILab/SLANTbrainSeg}.

\end{abstract}
\section{Introduction}
 Historically, multi-atlas segmentation (MAS) has been regarded as the de facto standard method on detailed whole brain segmentation ($>$ 100 anatomical regions) due to its high accuracy. Moreover, MAS only demands a small number of manually labeled examples (atlases)\cite{asman2014hierarchical}. Recently, deep convolutional neural networks (DCNN) have been applied to whole brain segmentation. To address the challenges of training a network on a small number of manually traced brains, patch-based DCNN methods have been proposed. de Brébisson et al., \cite{de2015deep} proposed to learn 2D and 3D patches as well as spatial information, which was extended to include 2.5D patches by BrainSegNet \cite{mehta2017brainsegnet}. Recently, DeepNAT \cite{wachinger2017deepnat} was proposed to perform hierarchical multi-task learning on 3D patches. Li et al., \cite{li2017compactness}  introduced the 3D patch-based HC Net for high resolution segmentation. From another perspective, Roy et al., \cite{roy2017error} proposed to use 2D fully convolutional network (FCN) to learn slice-wise image features by using auxiliary labels on initially unlabeled data. Although detailed cortical parcellations were not performed, Roy et al., revealed a promising direction on how to use initially unlabeled data to leverage training. With a large number of auxiliary labels, it is appealing to perform 3D FCN (e.g., 3D U-Net \cite{cciccek20163d}) on whole brain segmentation since it typically yields higher spatial consistency than 2D or patch-based methods. However, directly applying 3D FCN to whole brain segmentation (e.g., 1mm isotropic resolution) is restricted by the current graphics processing unit (GPU) memory. A common solution is to down sample the inputs, yet, the accuracy can be sacrificed.  
 
 In this paper, we propose the spatially localized atlas network tiles (SLANT) method for detailed whole brain segmentation (133 labels under BrainCOLOR protocol \cite{huo2016mapping}) by combining canonical medical image processing techniques with deep learning. SLANT distributes a set of independent 3D networks (“network tiles”) to cover overlapped sub-spaces in a standard MNI atlas space \cite{evans19933d}.

\begin{figure}[h]
\centering
\includegraphics[width=290pt]{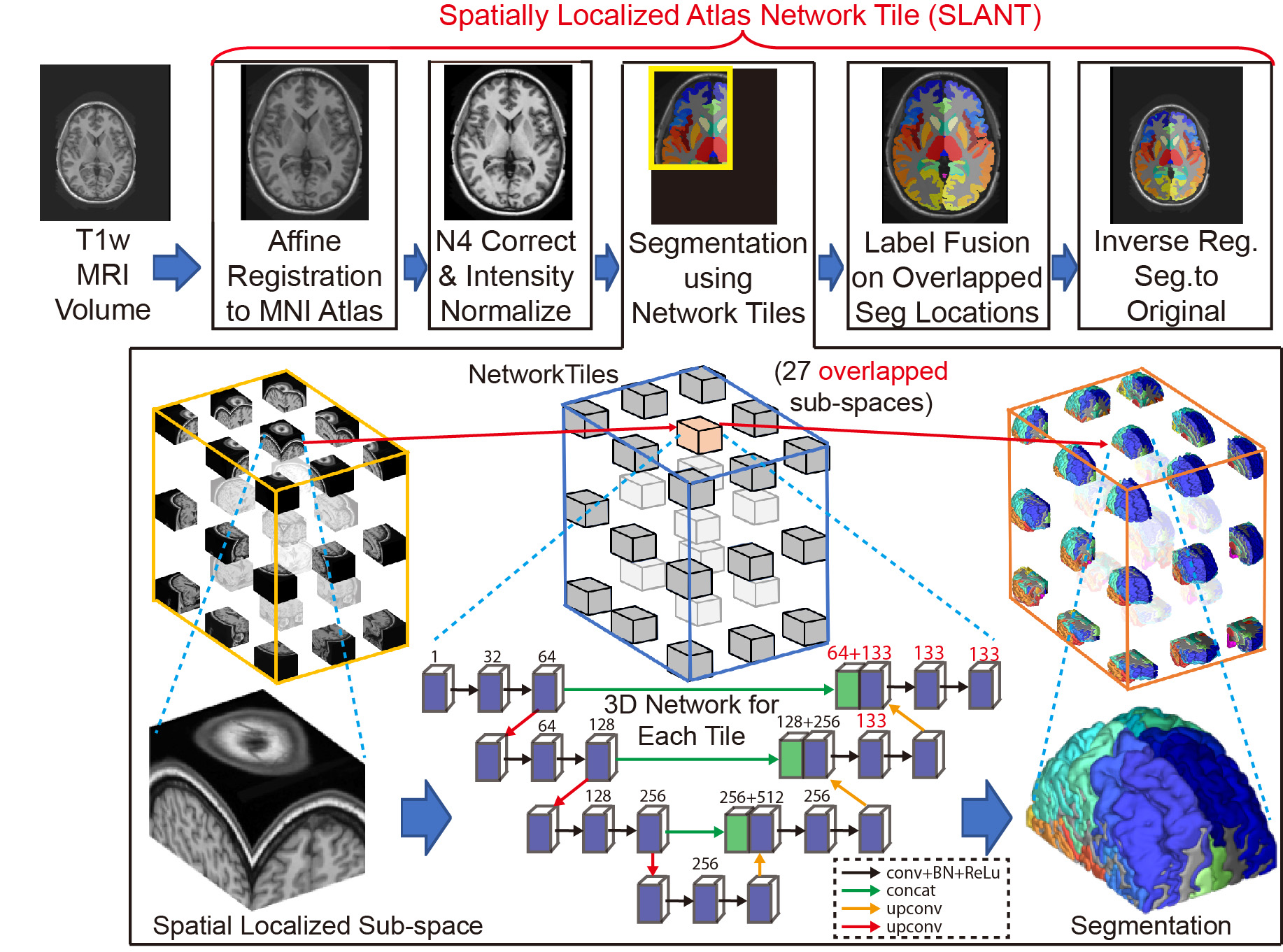}
\caption{The proposed SLANT-27 (27 network tiles) method is presented, which combines canonical medical image processing methods (registration, harmonization, label fusion) with 3D network tiles. 3D U-Net is used as each tile, whose deconvolutional channel numbers are modified to 133. The tiles are spatially overlapped in MNI space.} \label{fig1}
\end{figure}

 Then, majority vote label fusion was used to obtain final whole brain segmentation from the overlapped sub-spaces. To leverage learning performance on 133 labels with only 45 manually traced training data, auxiliary labels on 5111 initially unlabeled scans were created from non-local spatial STAPLE (NLSS) MAS \cite{asman2014hierarchical} for pre-training inspired by \cite{roy2017error}. 

\section{Methods}

\subsubsection{Registration and Intensity Harmonization:} Affine registration \cite{ourselin2001reconstructing} was employed to register all training and testing scans to MNI 305 space \cite{evans19933d} (Fig.~\ref{fig1}). Then, N4 bias field correction was deployed to reduce bias. To further harmonize the intensities on large-scale MRI, we introduced a regression-based intensity normalization method. First, we defined a gray-scale MRI volume (with N voxels) as a vector $I \in \mathbb{R}^{N\times1}$. $I$ was demeaned and normalized by standard deviation (std) to $I^{'}$. The intensities were harmonized by a pre-trained linear regression model on “sorted intensity”. The sorted intensity vector $V_s$ was calculated from $V_s=\textrm{sort}(I^{'}(mask>0))$, where "sort" rearrange intensities from largest to smallest, and  "$mask$" was a prior mask learned from a union operation from all atlases. To train the linear regression, mean sorted intensity vector  $\overline{V_s}$ was obtained by averaging $V_s$ from all atlases. . The coefficients were fitted between $V_{s}^{'}$ (from $I^{'}$) and $\overline{V_s}$, and intensity normalized image $\widehat{I^{'}}$ is obtained from fitted $\beta_1$ and $\beta_0$: $\overline{V_s} = \beta_{1} \cdot{} {V_{s}^{'}} + \beta_{0}$, and $\widehat{I^{'}} = \beta_{1} \cdot{} {I^{'}} + \beta_{0}$.

\begin{figure}
\centering
\includegraphics[width=310pt]{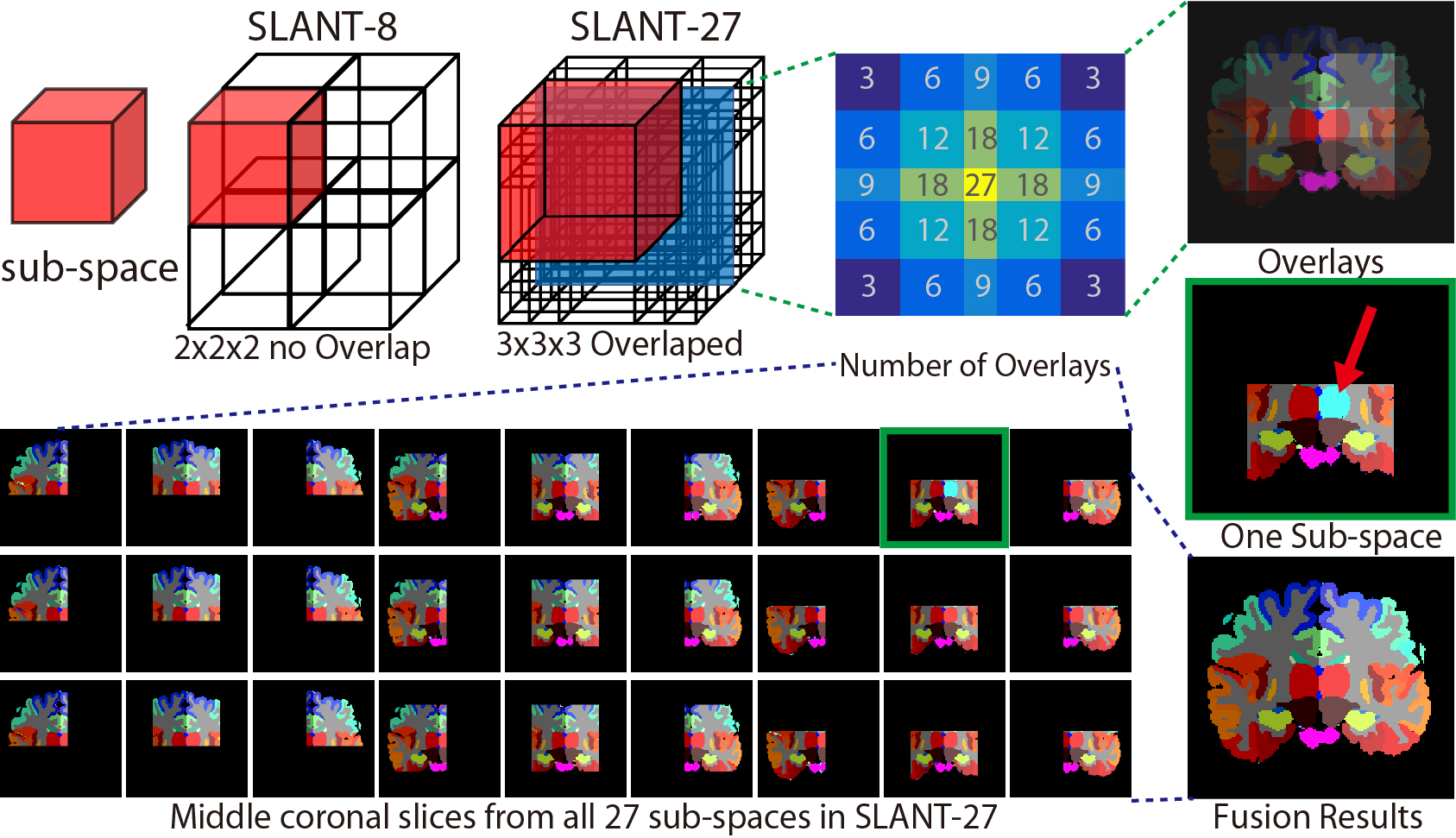}
\caption{SLANT-8 covered eight non-overlapped sub-spaces in MNI, while SLANT-27 covered 27 overlapped sub-spaces in MNI. Middle coronal slices from all 27 sub-spaces were visualized (lower left panel). The number of overlays, as well as sub-spaces’ overlays, were showed (upper right panel). The incorrect labels (red arrow) in one sub-space were corrected in final segmentation by performing majority vote label fusion.} \label{fig2}
\end{figure}

\subsubsection{Network Tiles:} After affine registration, all training brains were mapped to the same MNI atlas space ($172\times220\times156$ voxels with 1 mm isotropic resolution). We employed $k$ 3D U-Net as a network tiles to cover entire MNI space with/without overlaps (Fig.~\ref{fig2}). To be compatible with 133 labels, the number of channels of deconvolutional layers in each 3D U-Net were defined as 133 (Fig.~\ref{fig1}). Each sub-space $\psi_n$ was presented by one coordinate $(x_n,y_n,z_n)$ and sub-space size $(d_x,d_y,d_z)$, $n \in \{1,2,...,k\}$ as

\begin{equation}
\psi_n=[x_n:(x_n+d_x),y_n:(y_n+d_y),z_n:(z_n+d_z)]
\end{equation}
\noindent As showed in Fig.~\ref{fig2}, SLANT-8 covered the MNI space using eight U-Nets by covering $k=2\times2\times2=8$ non-overlapped sub-spaces. To improve spatial consistency at boundaries, SLANT-27 covered $k=3\times3\times3=27$ overlapped sub-spaces. 
\\
\\
\textbf{Label Fusion:} For SLANT-27, whose sub-spaces were overlapped, the label fusion method were employed to get a single segmentation from overlapped sub-spaces. Briefly, the k segmentations $\{S_1,S_2,...,S_k\}$ from network tiles were fused to achieve final segmentation $S_{MNI}$ in MNI space by performing majority vote:

\begin{equation}
S_{MNI}(i)=\mathop{\arg\min}_{l \in \{0,1,...,L-1\}}\frac{1}{k} \sum_{m=1}^{k} p(l|S_m,i)
\end{equation}
\noindent where $\{0,1,...,L-1\}$ represents $L$ possible labels for a given voxel $i \in \{1,2,...,N\}$. $p(l|S_m,i)=1$ if $S_m (i)=l$, while $p(l|S_m,i)=0$, otherwise. Then, the $S_{MNI}$ was registered to the original space by conducting another affine registration  \cite{ourselin2001reconstructing}. When using SLANT-8, whose sub-spaces were not overlapped, the native concatenation was applied rather than performing the label fusion.

\begin{figure}[h]
\centering
\includegraphics[width=\textwidth]{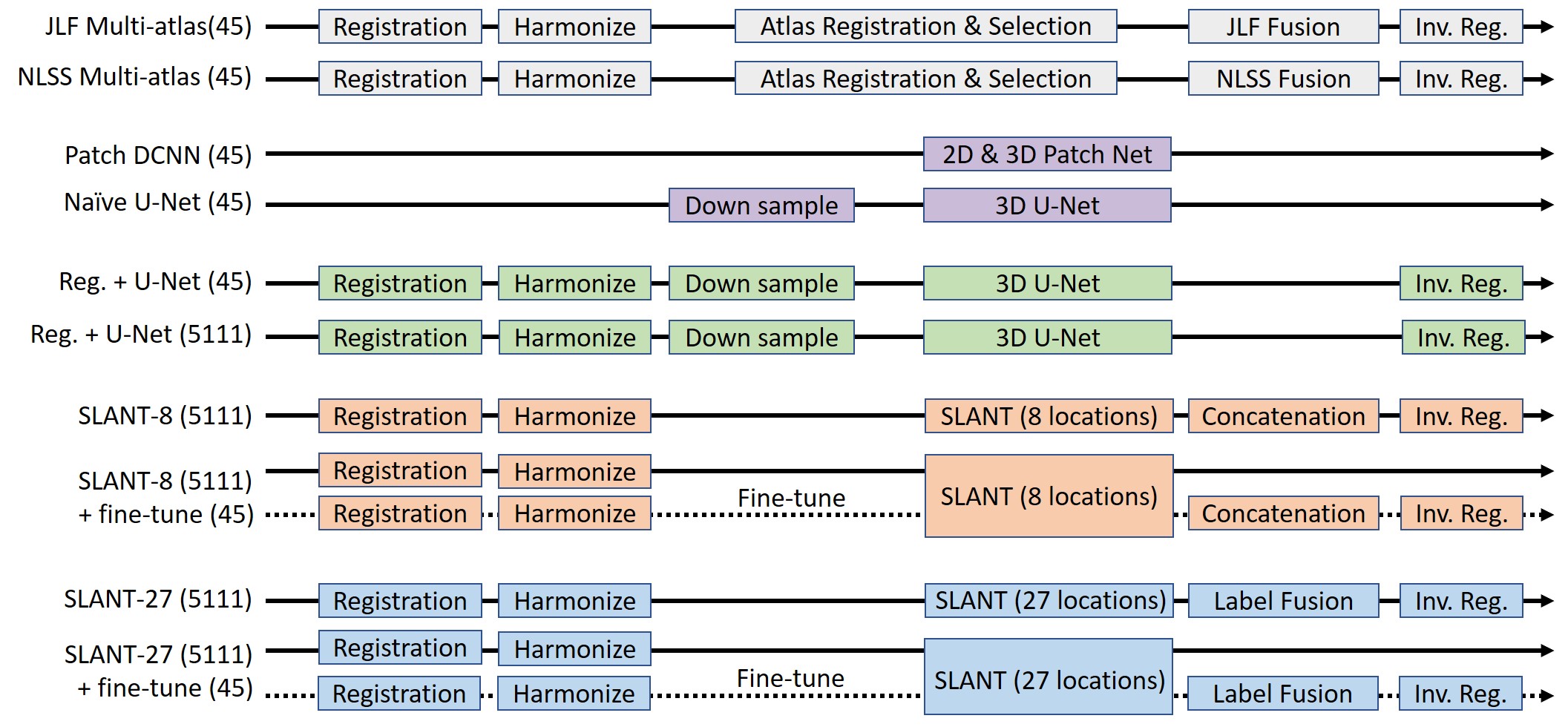}
\caption{This figure demonstrates the major components of different segmentation methods. “(45)” indicated the 45 OASIS manually traced images were used in training, while “(5111)” indicated the 5111 auxiliary label images were used in training.} \label{fig3}
\end{figure}

\subsubsection{Boost Learning on Unlabeled Data:} Similar to \cite{roy2017error}, the auxiliary labels on large-scale initially unlabeled MRI scans were obtained by performing existing segmentation tools. Briefly, MAS using hierarchical non-local spatial staple (NLSS) label fusion \cite{asman2014hierarchical} was performed on 5111 multi-site scans. Next, the large-scale auxiliary labels were used for pre-training. Then, the small-scale manually labeled training data were used for fine-tuning the network.

\section{Experiments}
\textbf{Training Cohort:} 45 T1-weighted (T1w) MRI scans from Open Access Series on Imaging Studies (OASIS) dataset \cite{marcus2007open} were manually labeled to 133 labels according to BrainCOLOR protocol \cite{huo2016mapping}. 5111 multi-site T1w MRI scans for auxiliary labels were achieved from night different projects (described in \cite{huo2016mapping}).

\noindent \textbf{Testing Cohort 1:} Five withheld T1w MRI scans from OASIS dataset with manual segmentation (BrainCOLOR protocol) were used for validation, which evaluates the performance of different methods on the same site testing data.

\noindent \textbf{Testing Cohort 2:} One T1 MRI scan from colin27 cohort \cite{collins1998design} with manual segmentation (BrainCOLOR protocol) was used for testing. This cohort evaluates the performance of different methods on a widely used standard template.

\noindent \textbf{Testing Cohort 3:} 13 T1 MRI scans from Child and Adolescent Neuro Development Initiative (CANDI) \cite{kennedy2012candishare} were used for testing. This cohort evaluates the performance of different methods on an independent population, whose age range (5-15 yrs.) was not covered by OASIS training cohort (18-96 yrs.).
\begin{figure}[h]
\centering
\includegraphics[width=310pt]{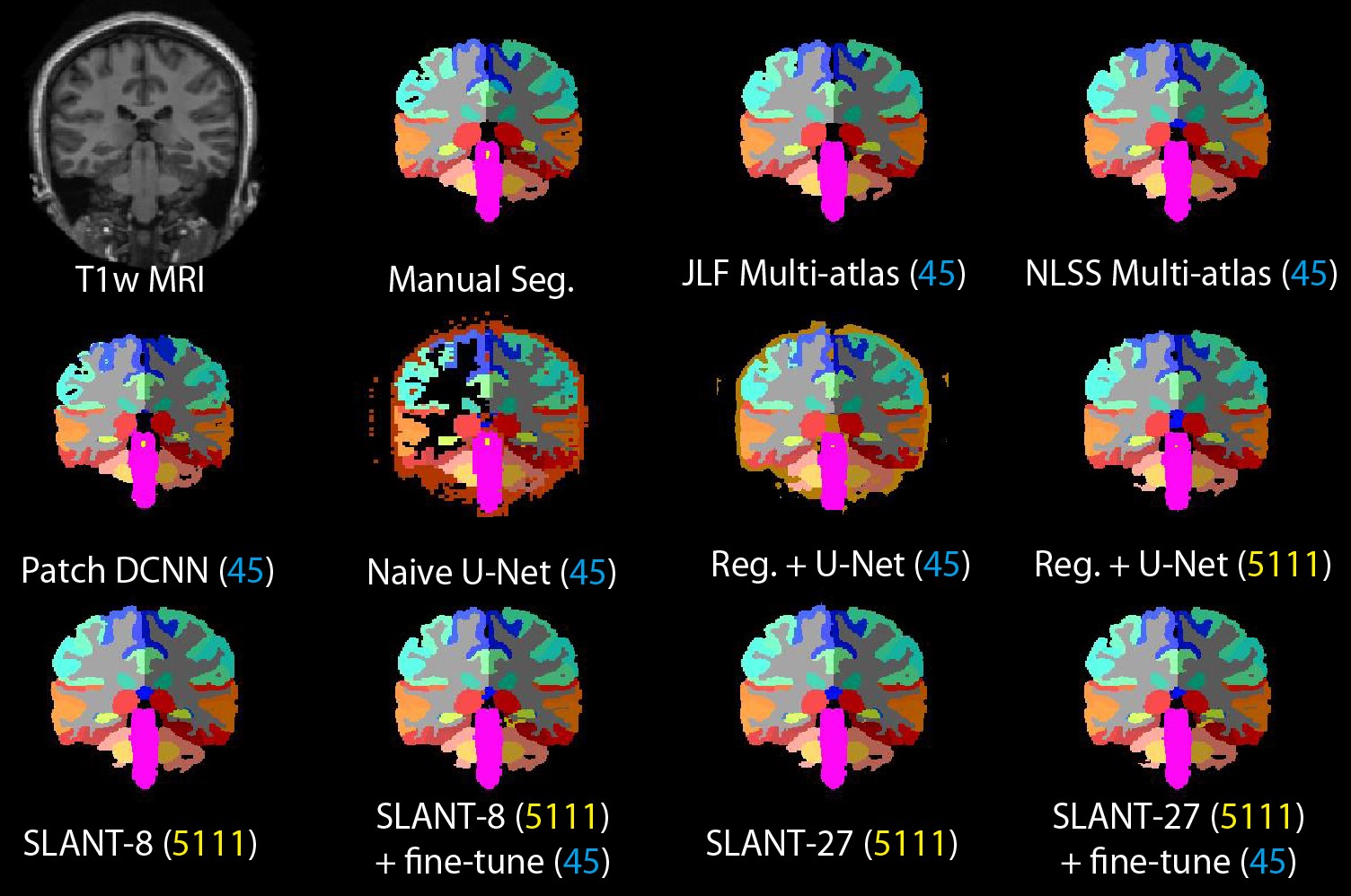}
\caption{Qualitative results of manual segmentation, MAS methods, patch-based DCNN method, U-Net approaches and proposed SLANT methods.} \label{fig4}
\end{figure}

\subsubsection{Experimental Design:} The experimental design is presented in Fig.~\ref{fig3}. First, two state-of-the-art multi-atlas label fusion methods, joint label fusion (JLF) \cite{wang2013multi} and non-local spatial staple (NLSS) \cite{asman2014hierarchical}, were used as baseline methods. The parameters were set the same as the papers’, which were optimized for whole brain segmentation. Next, patch-based network \cite{de2015deep} and naive 3D U-Net \cite{cciccek20163d} methods were used using their open-source implementations. By using affine registration as preprocessing, “Reg.+U-Net” was trained using 45 manually labeled scans and 5111 auxiliary labeled scans. Then, the proposed SLANT methods were evaluated on covering eight non-overlapped sub-spaces (“SLANT-8”) and 27 overlapped sub-spaces (“SLANT-27”), trained by 5111 auxiliary labeled scans. Last, 45 manually labeled scans were used to fine-tune the SLANT networks. 

For all 3D U-Net in baseline methods and SLANT networks, we used the same parameters with 3D batch size = 1, input resolution = $96\times128\times88$, input channel = 1, output channel = 133, optimizer = “Adam”, learning rate = 0.0001. The deep networks can fit into an NVIDIA Titan GPU with 12 GB memory. For all the training using 5111 scans, 6 epochs were trained ($\approx$ 24 training hours); while for all the training using 45 scans, 1000 epochs were trained to ensure the similar training batches as 5111 scans. For the fine-tuning using 45 scans, 30 epochs were trained. 

Results reported in this paper were from the epoch with best overall performance on OASIS cohort for each method, so that colin27 and CANDI were independent testing cohorts as external validation.

\begin{figure}[h]
\centering
\includegraphics[width=\textwidth]{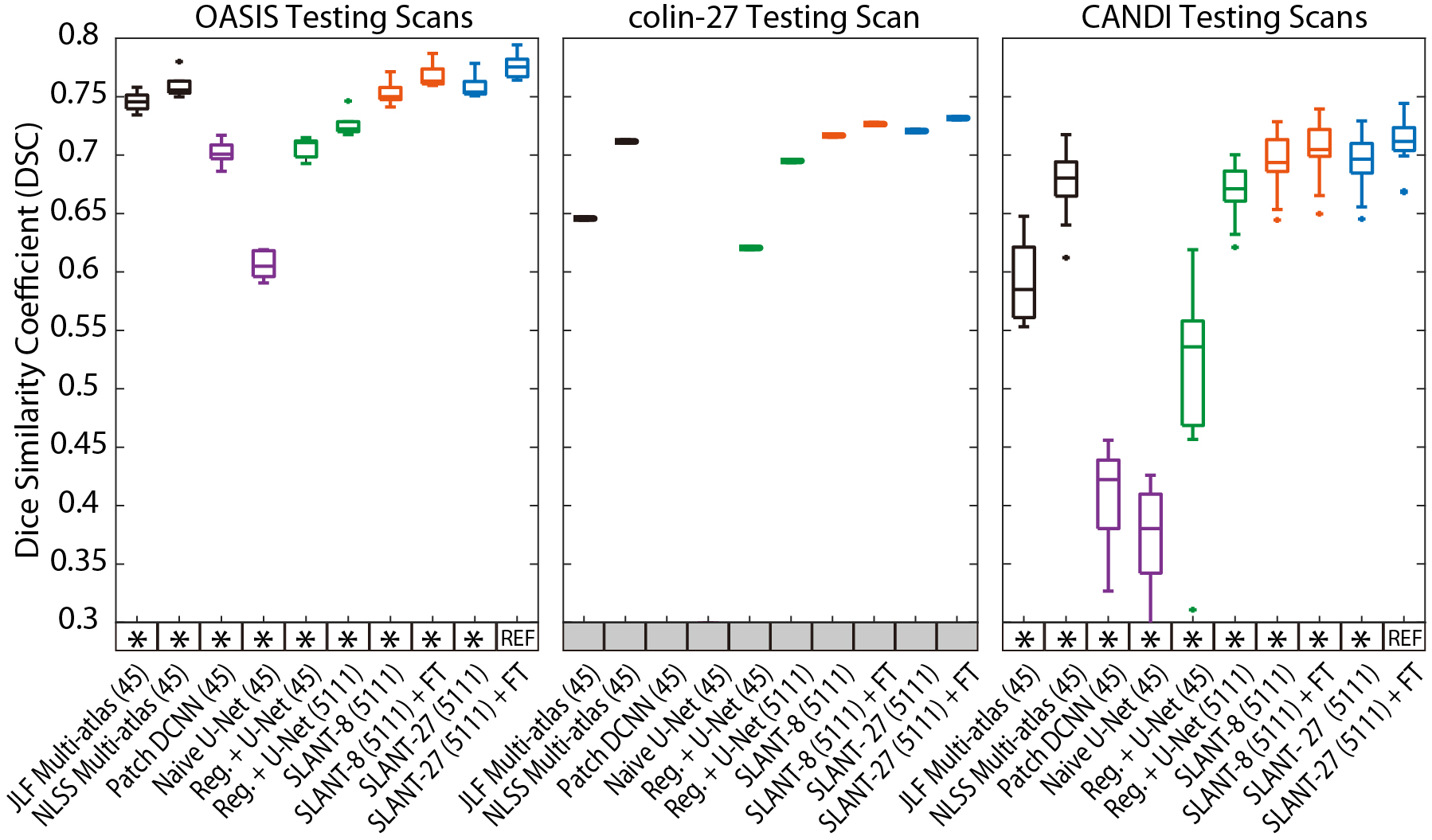}
\caption{From quantitative results, The propsed SLANT-27 using 5111 auxiliary labels for pretraining and fine-tuned (“FT”) by 45 manual labels achieved highest median  Dice similarity coefficient (DSC) values. The SLANT-27 was used as the reference method (“REF”) in statistical analysis. The significant difference to REF was marked with “*”.} \label{fig5}
\end{figure}

\section{Result}
The qualitative and quantitative results have been shown in Fig.~\ref{fig4} and \ref{fig5}. In Fig.~\ref{fig5}, ''45'' indicated the 45 OASIS manually traced images were used in training, while ''5111'' indicated the 5111 auxiliary label images were used in training. The mean Dice similarity coefficient (DSC) values on 132 anatomical labels (excluding background) between automatic segmentation methods with manual segmentation in original image space were showed as boxplots in Fig.~\ref{fig5}.  From the results, the affine registration (Reg. + U-Net) significantly leveraged the U-Net performance (compared with Naive U-Net). For the same network (Reg. + U-Net), results using 5111 auxiliary labeled scans achieved better performance than using 45 manual labeled scans. From Table \ref{table1}, the proposed SLANT-27 method with fine-tuning achieved superior performance on mean DSC across testing cohorts. All claims of statistical significance in this paper have been calculated using the Wilcoxon signed rank test for p $<$ 0.05.  

\section{Conclusion and Discussion}
In this study, we developed the SLANT method to combine the canonical medical image processing approaches with localized 3D FCN networks in MNI space. For the same network (Reg. + U-Net), results from 5111 auxiliary labeled scans achieved better performance than the results from 45 manual labeled scans. From Fig.~\ref{fig4} and \ref{fig5}, and Table~\ref{table1}, we demonstrate that our proposed strategy successfully takes advantages of historical efforts (registration, harmonization, and label-fusion) and consistently yields superior performance. Moreover, the proposed method requires $\approx$ 15 minutes, compared with $>$ 30 hours by MAS. Note that the 3D U-Net in the proposed SLANT can be replaced by other 3D segmentation networks, which might yield better performance. 

\begin{table}[]
\centering
\caption{Mean, std and median DSC values on three validation cohorts}
\label{table1}
\begin{tabular}{|c|c|c|c|c|c|c|}
\hline
\multirow{2}{*}{Methods} & \multirow{2}{*}{\begin{tabular}[c]{@{}c@{}}Training \\ Scan \#\end{tabular}} & \multicolumn{2}{c|}{OASIS Dataset} & Colin27 & \multicolumn{2}{c|}{CANDI Dataset} \\ \cline{3-7} 
&             & mean $\pm$ {\textup{std}}        & median        & DSC     & mean $\pm$ {\textup{std}}   & \begin{tabular}[c]{@{}c@{}}median\end{tabular} \\ \hline
JLF  \cite{wang2013multi}        & 45      & 0.746 $\pm$ 0.009 & 0.746 & 0.646 & 0.590 $\pm$ 0.033 & 0.585 \\ \hline
NLSS  \cite{asman2014hierarchical}        & 45      & 0.760 $\pm$ 0.012 & 0.756 & 0.712 & 0.677 $\pm$ 0.029 & 0.680 \\ \hline
Patch DCNN  \cite{de2015deep}  & 45      & 0.702 $\pm$ 0.011 & 0.701 & 0.012 & 0.409 $\pm$ 0.038 & 0.422 \\ \hline
Naive U-Net  \cite{cciccek20163d} & 45      & 0.606 $\pm$ 0.012 & 0.605 & 0.000 & 0.375 $\pm$ 0.043 & 0.380 \\ \hline
Reg. + U-Net         & 45      & 0.706 $\pm$ 0.009 & 0.711 & 0.621 & 0.514 $\pm$ 0.081 & 0.536 \\ \hline
Reg, + U-Net         & 5111    & 0.726 $\pm$ 0.012 & 0.722 & 0.695 & 0.669 $\pm$ 0.023 & 0.671 \\ \hline
SLANT-8              & 5111    & 0.753 $\pm$ 0.011 & 0.750 & 0.717 & 0.694 $\pm$ 0.024 & 0.694 \\ \hline
SLANT-8+FT         & 5111+45 & 0.768 $\pm$ 0.011 & 0.763 & 0.726 & 0.704 $\pm$ 0.025 & 0.705 \\ \hline
SLANT-27             & 5111    & 0.759 $\pm$ 0.011 & 0.754 & 0.721 & 0.694 $\pm$ 0.024 & 0.697 \\ \hline
SLANT-27+FT        & 5111+45 & \textbf{0.776} $\pm$ 0.012 & \textbf{0.775} & \textbf{0.732} & \textbf{0.711} $\pm$ 0.023 & \textbf{0.712} \\ \hline
\end{tabular}%
\end{table}

\subsubsection{Acknowledgments:} This research was supported by NSF CAREER 1452485, NIH R01EB017230, R21EY024036, R21NS064534, R01EB006136,  R03EB012461, R01NS095291, Intramural Research Program, National Institute on Aging, NIH.

%
%
%
\bibliographystyle{splncs04}
\bibliography{huo}
%





\end{document}